# Estimation and Early Prediction of Grip Force Based on sEMG Signals and Deep Recurrent Neural Networks

A. Ghorbani Siavashani[1] – A. Yousefi-Koma[2] – A. Vedadi[3]

## Abstract

Hands are used for communicating with the surrounding environment and have a complex structure that enables them to perform various tasks with their multiple degrees of freedom. Hand amputation can prevent a person from performing their daily activities. In that event, finding a suitable, fast, and reliable alternative for the missing limb can affect the lives of people who suffer from such conditions.

As the most important use of the hands is to grasp objects, the purpose of this study is to accurately predict gripping force from surface electromyography (sEMG) signals during a pinch-type grip. In that regard, gripping force and sEMG signals are derived from 10 healthy subjects. Results show that for this task, recurrent networks outperform non-recurrent ones, such as a fully connected multilayer perceptron (MLP) network. Gated recurrent unit (GRU) and long short-term memory (LSTM) networks can predict the gripping force with R-squared values of 0.994 and 0.992, respectively, and a prediction rate of over 1300 predictions per second. The predominant advantage of using such frameworks is that the gripping force can be predicted straight from preprocessed sEMG signals without any form of feature extraction, not to mention the ability to predict future force values using larger prediction horizons adequately. The methods presented in this study can be used in the myoelectric control of prosthetic hands or robotic grippers.

## Declarations

Funding: The work is based upon research funded by Iran National Science Foundation (INSF) under project No. 4013238.
Conflicts of interest/Competing interests: The Authors declare that there is no conflict of interest.
Availability of data and material and code availability: all data and codes are available at https://github.com/Atusa-gh/GrippingForcePrediction.

## 1 Introduction

One of the dramatic events that can happen to human beings is amputation, which can result from trauma, ischemia, infection, or malignancy. This tragic event dramatically decreases the quality of life of amputees. Moreover, hand amputation is one of the most common amputations. Upper limb amputations can be categorized into partial hand amputation, wrist disarticulation, trans-radial (below-elbow), trans-humeral (above-elbow), and shoulder disarticulation in accordance with the amputation level.

Prosthetic limbs have been the best solution for amputation, giving amputees the chance to regain their limb capabilities fully or partially. An active hand prosthesis is a device that replaces the missing hand and provides grasping functions. It is crucial to translate amputees' intentions accurately and quickly to develop a broadly useable prosthesis.

Many of the active prostheses are controlled by sEMG signals, which originate from muscles' electrophysiological and mechanical activations. Due to their ease of collection and noninvasiveness, sEMG signals are used for various applications, such as myoelectric control of prosthetics. Subsequently, deriving precise and fast methods for estimating movement parameters such as hand gestures, fingers or wrist angles, and gripping force from sEMG signals has been broadly investigated by researchers.

Recently, along with the successful development of artificial intelligence, more attention has been paid to neural network methods to estimate gripping variables. More specifically, the number of published papers associated with deep learning methods and sEMG signals has quadrupled between 2017 and 2018 [1]. Various types of neural networks have been proven to deliver an adequate level of error. As a result, many deep neural networks such as convolutional neural networks (CNN), recurrent neural networks (RNN), gated recurrent units (GRU), and long short-term memory (LSTM) have been deployed to estimate hand movements or gestures with considerable accuracy. For instance, upper-limb joint angles for multiple movements have been estimated from sEMG signals with the help of a deep neural network, and the deep network's performance has been proven to be much more precise than a simple multilayer perceptron network (MLP) [2]. Other than the framework of the neural networks, the EMG signal processing, and the feature extraction process play a significant role in the estimation [3]–[5]. Regarding the estimation of hand position variables such as fingers' angles, RNNs outperform non-recurrent neural networks with the same size in terms of mean absolute error. Moreover, adding more neurons and layers to the RNN offers only tiny improvements in estimation; therefore, preprocessing the input signals of the neural network has a more significant effect on its performance [6].

[1] Center of Advanced Systems and Technologies (CAST), School of Mechanical Engineering, University of Tehran, Tehran, Iran (atusa.ghorbani@ut.ac.ir)
[2] Center of Advanced Systems and Technologies (CAST), School of Mechanical Engineering, University of Tehran, Tehran, Iran (aykoma@ut.ac.ir)
[3] Center of Advanced Systems and Technologies (CAST), School of Mechanical Engineering, University of Tehran, Tehran, Iran (amirhosein.vedadi@ut.ac.ir)

Nowadays, position control is implemented in the majority of robots performing position-strict tasks all around the world. Nevertheless, this is not well-suited for close human interactions due to safety issues. On the contrary, force control is much more effective for such applications, especially rehabilitation. A compliant behavior of a robotic hand can provide safe interactions between the amputee and the prosthetic device, which prohibits secondary injuries to the user and damages to objects being tackled by the device. However, very few prosthetic limbs or grippers function via force control [7].

Having great applications in rehabilitation and biomedical engineering, estimating muscular forces of both upper and lower limbs has also been studied. To exemplify, both sEMG signals and inverse dynamic analysis have been used to estimate upper-limb muscular forces [8]. Also, multiple muscle models have been deployed and compared for predicting muscular forces of gastrocnemius medialis, gastrocnemius lateralis, soleus, and tibialis anterior muscles from sEMG signals [9].

In addition, estimating the gripping force or fingers' force values have been studied to some extent. For example, by extracting the main time-domain features of the sEMG signals, the force of the power-type gripping has been estimated with reasonable precision via a general regression neural network (GRNN) [10]. Also, a three domains fuzzy wavelet neural network (TDFWNN) algorithm alongside a conventional radial basis function neural network (RBFNN) method has been utilized to estimate power-type gripping force with an acceptable precision [11]. Regarding other machine learning algorithms, finger forces during a power-type grip have been successfully estimated via a gradient boosting machine (LightGBM) model [12].

Moreover, using deep neural networks (DNNs) as well as dimension reduction methods such as principal component analysis (PCA) in estimating the gripping force of the pinch-type grip has shown promising results for myoelectric control of a prosthetic hand [13]. Besides backpropagation neural networks, other machine learning methods such as gene expression programming algorithm (GEP) have also shown remarkable results in estimating gripping force [14]. Using non-negative matrix factorization (NMF) based input signal extraction for neural networks such as CNN and LSTM combined with strong signal processing instead of the traditional feature extraction methods can also result in precise estimations [15].

In this paper, to achieve a quick and precise prosthetic hand, the gripping force of the three-finger pinch mode, which is one of the most commonly used gripping gestures, is estimated straight from the filtered and normalized sEMG signals using a variety of neural networks. Furthermore, the performance of simple RNN, LSTM, GRU, and MLP networks for this task are compared.

## 2 Material and methods

At the beginning of this section, experiment protocols are defined. Then, a schematic figure of the whole predicting process is presented, followed by a description of the signal processing of sEMG and force signals. Lastly, all implemented neural networks are presented.

### 2.1 Experiment protocols

A test setup based on a three-finger pinch mode grip is designed to derive an adequate force-EMG data set. In this setup, EMG signals of the forearm muscles and the compressive force of the three-finger pinch grip are simultaneously measured.

An ATI 6-axis force/torque sensor from the mini45 series is used to measure the gripping force. This sensor is capable of recording all six components of force and torque. However, only one component of the force created by applying pressure on the z-axis of the sensor is sufficient for estimating the pinching gripping force.

A Myo armband bought from the Thalmic Labs is used to measure the sEMG signals. This armband has 8 EMG channels that can be wrapped around an arm and measure and record electrical impulses given off by the hand's muscles via the sEMG sensors. By digesting these pulses, the armband itself can control an electronic device based on the hand's gestures. The raw EMG signals of all eight sensors can also be derived through the Bluetooth port via Python. The Myo armband is put in the middle of the forearm-length area of the hand. Between three different electrode positions in the superficial forearm muscles' positions such as the extensor digitorum muscle, Flexor digitorum superficialis muscle, and Palmaris longus muscle, setting the armband in the middle of the forearm produces the best results in finger movement classification [16]. This is the ground reason why the Myo armband has been placed in the upper middle of the forearm muscle during the testing of this research.

Since the arm position has proved to affect discrimination between upper limb motion classes when using both surface and intramuscular EMG [17] and flexion of the wrist also influences the grip force [18], a fixed state is considered during all tests to minimize these disturbances. The subject's hand and forearm positions on the table are such that the elbow and wrist angles are 90 and 0 degrees, respectively. The process of one test and both the force sensor and Myo armband are shown in Fig. 1. In each test, the subject first applies pressure on the F/T sensor, decreases the pressure, and repeats this action for 30 seconds. Since the change of the EMG signals can be visually seen during applying pressure on the sensor, the EMG signals are plotted live during each test to monitor the whole experiment. There is a 60-second pause between each test of one subject and each subject repeats the test three times, making it a total of 3 minutes and 30 seconds for each subject's test. It should be noted that an exact measurement of the velocity of the grip was deliberately not considered because, ideally, the user of the prosthesis will want to perform the gripping with different velocities. Consequently, it's more effective to include different gripping velocities in the training data as well. The data set in this project is derived from 10 subjects with an average age of 23.8 years old (9 males and one female). Participants were chosen from healthy students of the University of Tehran and they all were fully informed about the nature of the study and its details. University's ethics committee has approved this experiment.

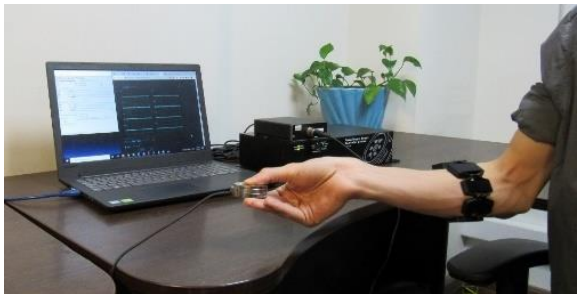

**Fig. 1** The experimental protocol

It should be noted that the F/T sensor reports the measured force in counts which can be converted to Newton with a simple multiplication according to the sensor's datasheet. The maximum force measured in the tests is in the range of 30-70 N, according to the maximum pinch-type grip force reported, which is 82 N for males in their twenties [19]. The signal derived from one test is shown in Fig. 2.

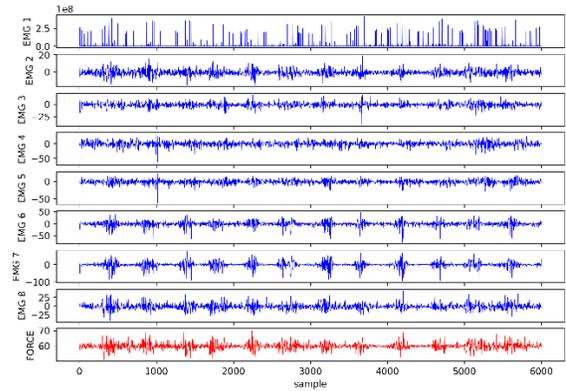

**Fig. 2** Raw sEMG and force signals of one subject (30 seconds)

## 2. 2 Signal preprocessing

A schematic of the whole force prediction process consisting of data collection, signal processing, network training, and performance analysis is shown in Fig. 3. In summary, at the beginning of the process, the raw sEMG signals of the Myo armband's eight channels and the gripping force are collected simultaneously using Python 3.8 via Bluetooth and serial port, respectively. Subsequently, all data is preprocessed to achieve a higher prediction accuracy. A band-pass filter is designed for each data type to limit the bandwidth of data signals. In addition, a notch filter is also used to attenuate signals over a narrow range of frequencies while leaving the signal at other frequencies unaltered to eliminate the effects of transmitting data through wires and common noises. The cut-off frequencies of the implemented filters are carefully chosen with regards to both filter results and the frequency of the muscle activation, which has been observed from 20 Hz to 60 Hz [20]. Muscle activations are adequately recorded and preserved during filtering with the Myo armband sampling frequency of 200 Hz. The F/T sensor has a much higher sampling rate. Therefore, to make the sampling frequencies of both sensors compatible with each other, both sEMG and force data are measured simultaneously using one code with a frequency of 200 Hz.

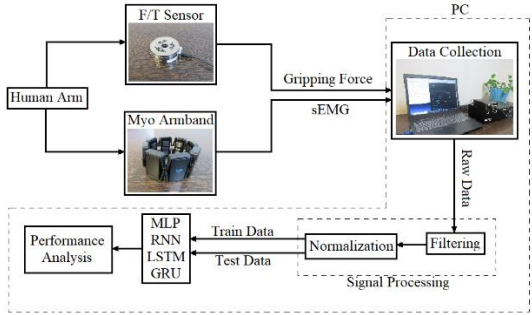

**Fig. 3** Schematic diagram of the whole force prediction process

In the next step of the process, the filtered data is normalized. This step aims to change the values of dataset columns to a standard scale without distorting differences in the ranges of values. In other words, the data is normalized so that all variables are in the same range. It should be noted that since each person can implement a different maximum force level, the data obtained from all subjects is first merged and then all data is normalized. This way, the algorithm is able to distinguish the different force levels that different subjects implement and, as a result, is more adjustable for different people. The Min-Max scale normalization, which is very common in neural networks, is used in this research. This method puts the data between 0 and 1. Fig. 4 shows one sample of filtered gripping force signals as well as filtered and normalized sEMG signals.

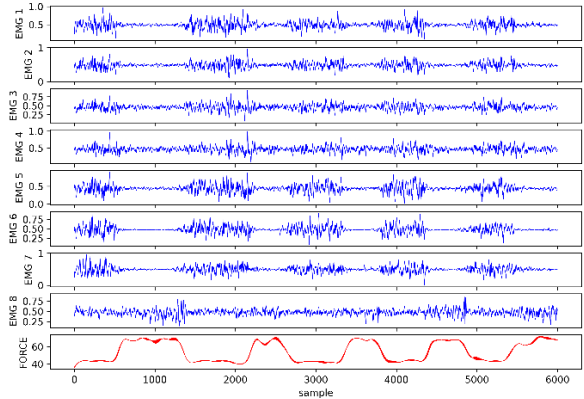

**Fig. 4** Filtered gripping force and filtered and normalized sEMG signals

### 2. 3    Neural networks in force prediction

The human hand is a dynamic system; therefore, the collected data in this research is a time series, meaning the data received at a specific time is dependent on the previously collected data. Recurrent neural networks can learn long-term features from sequential and time-series data sets. In other words, RNNs can learn past complex data for long periods because of the non-linear structures of their layers [21]. This is the principal reason why recurrent neural networks have more precise predictions compared with non-recurrent ones. To prove this point, a fully connected MLP network is compared with a simple RNN. More complex recurrent neural networks such as GRU and LSTM are also implemented to achieve higher precisions. In the following, each of the networks mentioned above is described in sufficient detail.

MLP network is widely used for various classification and regression applications. The fully connected MLP with two hidden layers implemented in this study is shown in Fig. 5. The input data of the MLP must be a vector of only one column, while recurrent neural networks can receive matrixes as inputs. The input data from each time step is stacked to form a vector to train the MLP network with the built dataset. The first and second hidden layers in the MLP have 200 and 80 neurons, respectively.

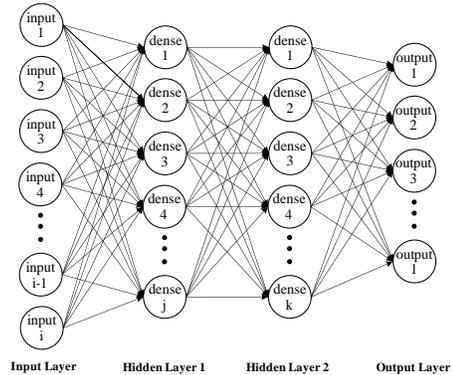

**Fig. 5** Fully connected MLP Neural network

The first recurrent neural network implemented in this study is a simple RNN. Input data goes to the recurrent layer, and the recurrent layer consists of an input layer, a hidden layer with 50 neurons, and an output layer. The structure of this layer is shown in Fig. 6. The output layer of the recurrent layer is connected to a fully connected layer with 100 neurons and then to the main output layer.

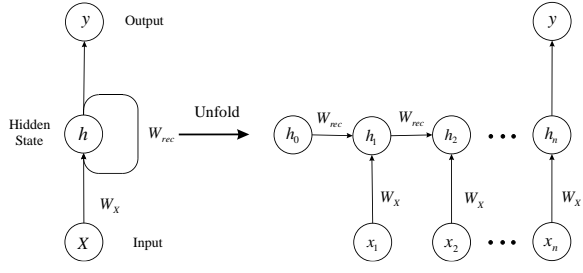

**Fig. 6** Structure of the recurrent layer

The output of the hidden layer is called a hidden state ($h_i$), and this output is affected by both the input data and the hidden state of the previous time step. Also, $W_x$ and $W_{rec}$ are weight matrices between the current input and the previous hidden state, which are updated in the training process. The hidden state in each layer is obtained according to Eq. (1). The function, f, in this equation is a Rectified Linear Unit (ReLU) function.

$$h_i = f(W_{rec} h_{i-1} + W_x x_i) \quad (1)$$

One major problem of the simple RNN network is the vanishing gradient, limiting its ability to learn long-term dependencies. LSTM networks solve this issue. This network was first introduced by Hochreiter and Schmidhuber [22]. In addition to the hidden state, the cell state is also used to transfer long-term dependencies in this network. Fig. 7 shows a view of the hidden unit in the LSTM layer. $C$ and $h$ are cell state and hidden state, respectively. Also, subscripts of n and n-1 represent current and previous step times.

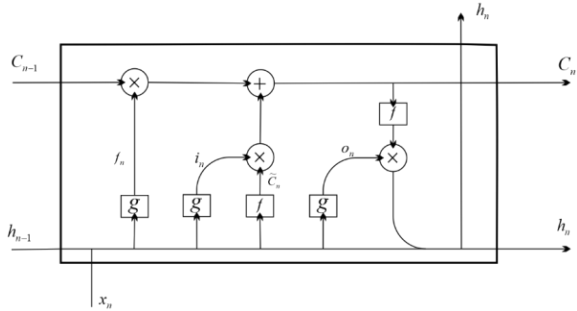

**Fig. 7** Hidden unit of LSTM layer

In each step time, long-term dependencies are applied to the cell state through various gates. These gates are used to add or remove information to the information stream in each network unit, whose weights and biases are updated in the training process. Also, the cell state affects the hidden state. Hidden state and cell state are obtained from Eq. (2), in which $f$ and $g$ functions are ReLU and sigmoid functions, respectively. × is the element-wise multiplication of vectors. Also, in each equation, $W$ and $b$ are the related weight matrix and bias vectors. $f_n$, $i_n$ and $o_n$ represent forget gate, input gate, and output gate, respectively. Moreover, $\tilde{C}_n$ is an intermediate gate for calculating the cell state. In the forget gate, it is decided which cell state information should be deleted. Furthermore, new information is added to the cell state in the input gate, and finally, the output gate is used to create the hidden and output states.

$$\begin{aligned} f_n &= g(W_f[h_{n-1}, x_n] + b_f) \\ i_n &= g(W_i[h_{n-1}, x_n] + b_i) \\ \tilde{C}_n &= f(W_c[h_{n-1}, x_n] + b_c) \\ C_n &= f_n \times C_{n-1} + i_n \times \tilde{C}_n \\ o_n &= g(W_o[h_{n-1}, x_n] + b_o) \\ h_n &= o_n \times f(C_n) \end{aligned} \quad (2)$$

GRU is a simpler version of LSTM networks, from which the cell state has been removed, and it tries to send the information of long-term dependencies by the hidden state only [23]. Also, some gates have been merged. Fig. 8 shows the hidden unit in the GRU network of this research.

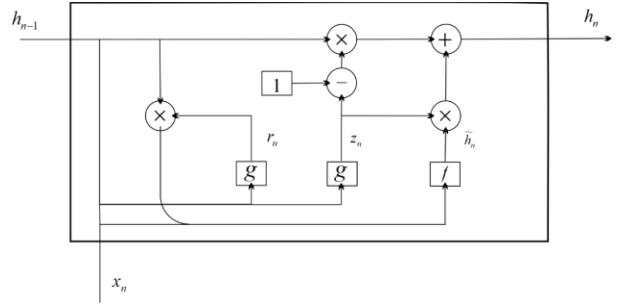

**Fig. 8** Hidden unit of GRU layer

In this network, a hidden state is obtained with the help of Eqs (3). Similar to Eqs (2), $f$ and $g$ are ReLU and sigmoid functions, respectively, and $W$ and $b$ are related weight matrix and the bias vector. In Eqs (3), $z_n$, $r_n$, and $\tilde{h}_n$ are all intermediate gates for finding a new hidden state $h_n$.

$$z_n = f(W_z[h_{n-1}, x_n] + b_z)$$
$$r_n = g(W_r[h_{n-1}, x_n] + b_r)$$
$$\tilde{h}_n = f(W[r_n \times h_{n-1}, x_n] + b) \quad (3)$$
$$h_n = (1 - z_n) \times h_{n-1} + z_n \times \tilde{h}_n$$

## 3 Results and discussions

To compare the results of the networks, they should be trained under the same conditions, and the same parameters must be used to train them. These parameters are specified in Table I.

**Table I** Networks' parameters

| Parameter | Value |
|---|---|
| Loss Function | Mean Squared Error (MSE) |
| Optimizer | Adam |
| Epochs | 60 |
| Batch Size | 32 |
| Evaluation Data | 15% |

Adam is an optimization method that has replaced the stochastic gradient descent method and has shown good performance in training neural networks [24]. All networks are compared using various performance criteria, including mean squared error (MSE), $R^2$ score, and test time. The first two mentioned criteria are both a type of metrics used to evaluate the regression models' performance, especially statistical ones. Their significant difference is that MSE is defined based on whether the data is scaled and captures the residual error; whereas, $R^2$ represents the fraction of variance of the response variable captured by the regression model. These two criteria are calculated based on the formulations in Eqs (4) and (5). $\mu_F$ in the latter is the mean of the actual forces.

Test time is the amount of time each neural network needed to predict the force values of the test data.

$$MSE = \frac{1}{n} \sum_{i=1}^{n} (F_i^{actual} - F_i^{predited})^2 \quad (4)$$

$$R^2 = 1 - \frac{\sum_{i=1}^{n}(F_i^{actual} - F_i^{predited})^2}{\sum_{i=1}^{n}(F_i^{actual} - \mu_F)^2} \quad (5)$$

Each network was trained and tested at different prediction horizons. The prediction horizon refers to the number of future force values that the network predicts. A higher prediction horizon provides information about further future, which can be used to make more accurate decisions in advance, provided that the error of these predictions does not increase. After training, the performance of the networks with the test data, which is 15% of the total data for different prediction horizons, is studied. A comparison between all implemented networks can be seen in Table II, which depicts the precise MSE of the normalized force values, $R^2$ score of the predictions, and test time for normalized test data at different prediction horizons.

As seen in Table II, in general, as the value of the prediction horizon increases, the MSE of each network increases, and conversely, $R^2$ decreases. Using larger prediction horizons has many practical benefits. For instance, it allows us to be aware of future events faster and have more time for control calculations. Moreover, we can use control methods such as the model predictive control scheme by predicting the output in the next few steps.

**Table II** Comparison of neural networks' performance for different prediction horizons

| | prediction horizon | MSE | $R^2$ | Test time (s) |
|---|---|---|---|---|
| MLP | 1 | 0.0016848 | 0.9788782 | 1.512 |
| Simple RNN | | 0.0011560 | 0.9855077 | 5.592 |
| GRU | | 0.0004762 | 0.9940306 | 10.808 |
| LSTM | | 0.0006643 | 0.9916720 | 10.509 |
| MLP | 3 | 0.0020688 | 0.9740624 | 02.730 |
| Simple RNN | | 0.0016571 | 0.9792235 | 10.851 |
| GRU | | 0.0010200 | 0.9872116 | 10.815 |
| LSTM | | 0.0014009 | 0.9824362 | 21.182 |
| MLP | 5 | 0.0027825 | 0.9651110 | 2.759 |
| Simple RNN | | 0.0033593 | 0.9578787 | 10.773 |
| GRU | | 0.0028747 | 0.9639549 | 10.875 |
| LSTM | | 0.0036569 | 0.9541473 | 21.142 |

Among the results with a prediction horizon of one, the most accurate method is the GRU recurrent neural network which can predict gripping force with an $R^2$ error of 0.9940306, closely followed by the LSTM network with an $R^2$ error of 0.9916720. The simple RNN network is placed third with an $R^2$ of 0.9855077, while the MLP network is placed last with an $R^2$ error of 0.9788782. Conversely, GRU and LSTM have rather long prediction times, predicting the test data in 10.808 and 10.509 seconds, respectively; whereas, MLP has the quickest prediction rate by predicting the test data in only 1.512 seconds. Lastly, the simple RNN takes 5.592 seconds to predict the force values of the test data. The difference between the networks' prediction accuracy and test time is due to their different topologies and hyperparameters. For instance, recurrent networks, especially LSTM and GRU, can outperform MLP because their topologies are more capable of preserving information from previous data points in an effective manner. Moreover, the low test time of MLP is due to its simplicity and low computational cost. Comparing LSTM and GRU in higher prediction horizons, GRU can predict the gripping force with higher accuracy. GRU also has a much lower test time, which means it would be a strongly preferable choice for Internet of Medical Things (IoMT) technologies or embedded systems. For further elaboration, Fig. 9 shows the performance of all the implemented neural networks in this study by depicting each network's predicted gripping force values with prediction horizons of 1, 3, and 5 and the actual gripping force values for 15 seconds of the normalized test data. This figure shows that all recurrent networks can precisely estimate the grip force with a prediction horizon of one. Early prediction of force, implemented by using higher prediction horizons, is also done with adequate error levels. This figure also demonstrates that the effect of the prediction horizon on MLP's performance is much smaller than that of other recurrent networks.

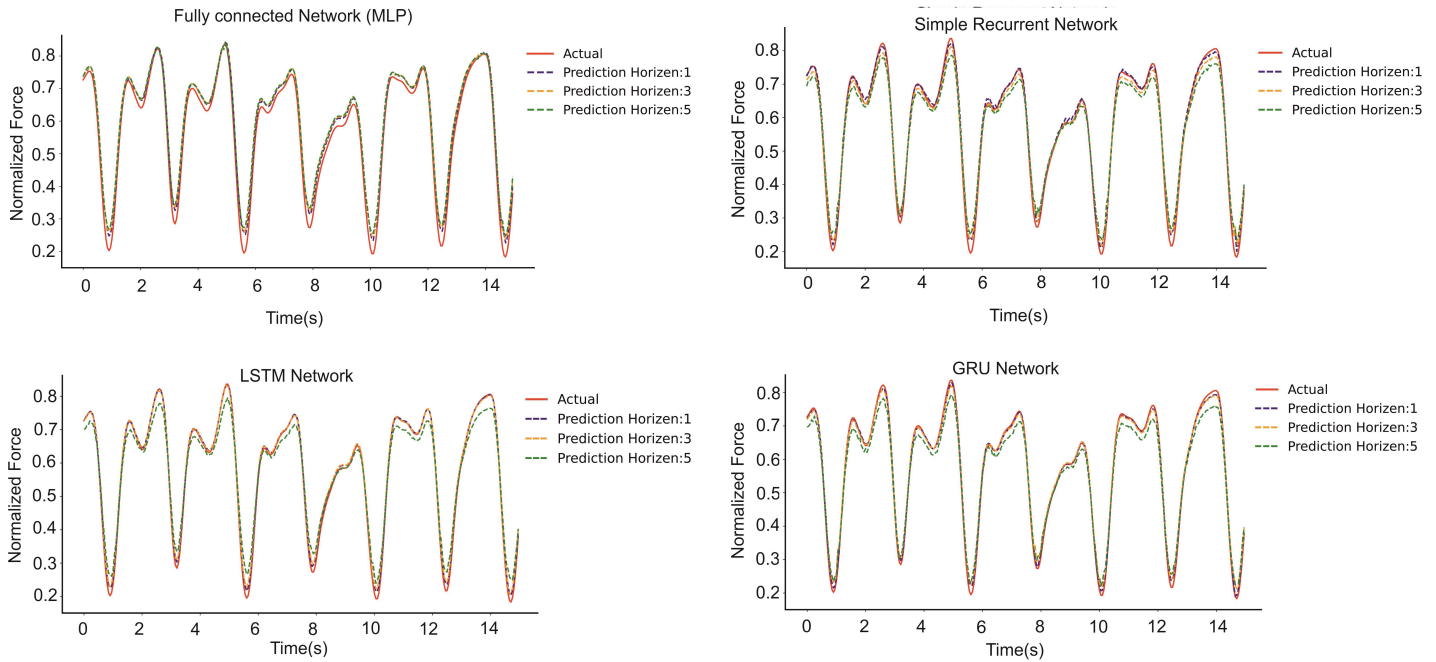

**Fig. 9** Networks' performance with prediction horizons of 1, 3, and 5 for 15 seconds: MLP (top left), simple RNN (top right), LSTM (bottom left), and GRU (bottom right)

Furthermore, Fig. 10 entails the boxplots of squared error distribution for each neural network with different prediction horizons. Recurrent networks outperform the non-recurrent network because they have much smaller maximum squared error values for each prediction horizon. The GRU has the smallest interquartile range, median and maximum error value in all cases, especially with a prediction horizon of 5, which depicts the superiority of the GRU network. Moreover, increasing the prediction horizon, which enables us to predict the gripping force in advance, decreases the prediction accuracy. This can be seen in the boxplots of each network.

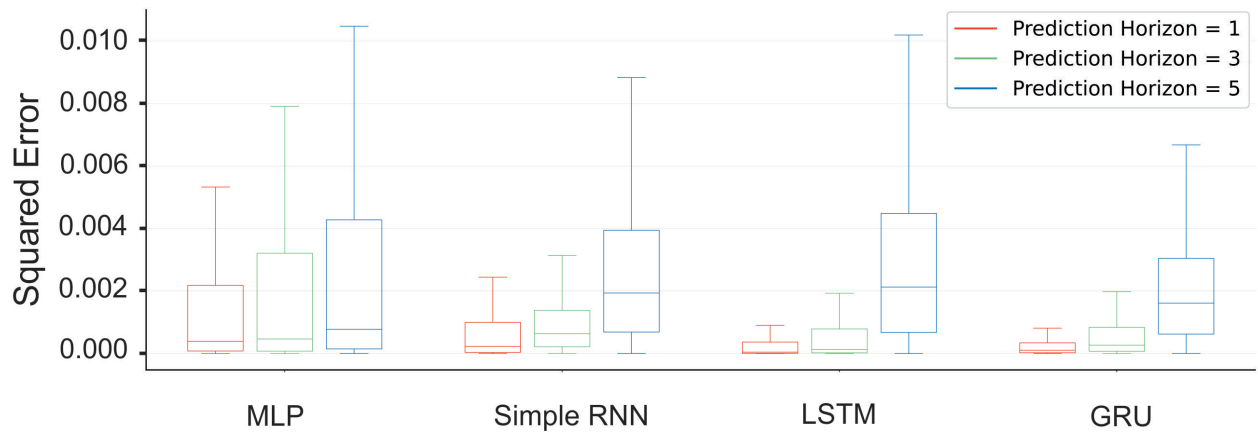

**Fig. 10** Distribution of Squared Error for different networks and prediction horizons

## 4 Conclusions

In this paper, we proposed various neural networks for predicting the gripping force of the pinch-type grip of a fixed arm position. More specifically, a fully connected MLP, a simple RNN, a GRU, and an LSTM network are trained for this task. Using the frameworks implemented in this study, the gripping force is quickly and highly accurately predicted straight from processed raw sEMG signals and without any form of feature extraction. The performance of these four networks is compared, and hence, the superiority of the GRU neural network is demonstrated. The neural networks proposed in this paper can be utilized for controlling myoelectric-based grippers and prosthetic hands. For future work, the force and EMG data of different arm positions can be obtained to re-train all or parts of the proposed ANNs in this paper via transfer learning methods, so that the network's performance is modified for a wider range of arm positions. This can lead to developing a fast and robust gripper or hand prostheses capable of controlling gripping force in numerous gripping settings.